\documentclass[letterpaper, 10 pt, conference]{ieeeconf}  % Comment this line out if you need a4paper

\IEEEoverridecommandlockouts                              % This command is only needed if 
\usepackage[font=small,labelfont=bf]{caption}
\usepackage[pdftex]{graphicx}
\usepackage{graphics} % for pdf, bitmapped graphics files
\usepackage{float} % H parameter in figures to have absolute control of positioning
\usepackage{hyperref} % for links to be clickable
\hypersetup{
  colorlinks   = true, %Colours links instead of ugly boxes
  urlcolor     = blue, %Colour for external hyperlinks
  linkcolor    = blue, %Colour of internal links
}
\usepackage{epsfig} % for postscript graphics files
\usepackage{amsmath} % assumes amsmath package installed
\usepackage{amssymb}  % assumes amsmath package installed
\usepackage{enumerate}
\usepackage{amsfonts}
\usepackage{subfig}
\usepackage{epstopdf}
\usepackage{booktabs}
\usepackage{array}
\usepackage{xcolor}
\usepackage[italian]{babel}
\usepackage{algorithm}
\usepackage{algpseudocode}

\usepackage[utf8]{inputenc}
\usepackage{verbatim}
\usepackage[active]{srcltx}
\usepackage{url}
\usepackage[T1]{fontenc}
\usepackage{mathtools} 
\usepackage{subfig}
\usepackage{cite}
\usepackage{microtype} % for local kerning: \textls 
\usepackage{bm}
\usepackage{dsfont}

\title{\LARGE \bf
Control Barrier Functions Solved with Hierarchical Quadratic Programming for Safe Physical Human-Robot Interaction
}

\author{Rui Luo, Jonas Mariager Jakobsen, Wesley Roozing, Federico Califano, Cheng Fang% <-this % stops a space
}

\begin{document}

\maketitle
\thispagestyle{empty}
\pagestyle{empty}

%%%%%%%%%%%%%%%%%%%%%%%%%%%%%%%%%%%%%%%%%%%%%%%%%%%%%%%%%%%%%%%%%%%%%%%%%%%%%%%%
\begin{abstract}

Physical human-robot interaction offers the potential to leverage human intelligence and robot physical capabilities to enable a range of exciting applications, e.g., collaborative robots for rehabilitation. Safety is critical for the successful deployment of this kind of robotic system. In recent years, Control Barrier Function (CBF) has emerged as an effective approach to enforce safety guarantees, which has been widely applied in various applications, from adaptive cruise control to navigation of legged robots. CBFs can be solved in a Quadratic Programming (QP) problem, which can include many CBF-formulated tasks. To manage a large number of safety tasks, a hierarchical CBF has been used to allow hierarchical relaxation of safety tasks to ensure the feasibility of a solution in the presence of conflicting tasks. In this work, we propose to use a CBF-based Hierarchical Quadratic Programming (HQP) framework in physical human-robot interaction to allow us to design both performance tasks (e.g., preserve the desired behavior at the human-robot interaction point) and safety tasks at any level of a hierarchy to balance the safety and the performance in a more flexible way. Extensive experiments were carried out on a real redundant robot to validate the effectiveness, flexibility, and generality of this approach.

\end{abstract}

\section{Introduction}

% \textcolor{blue}{a suggested text flow: \newline
% -> safety is important for the deployment of pHRI \newline
% -> CBF has become one of the mainstream approaches for safety-critical applications because of some benefits \newline
% -> there are many applications of CBF in safety-critical applications like collision avoidance, ..., but very few in safe pHRI. \newline
% -> In the previous applications, most of the researchers formulate all the safety tasks with CBFs as inequality constraints in a single QP problem, which is fine. \newline
% -> However, when we try to apply the CBF framework to the safe pHRI with one qp for multiple different safety tasks, we have problems because we have some interaction tasks that we want to preserve as much as we can, that is, can we prioritize the modification of the null-space joint torque component over that of the joint torque component that would affect the desired interaction behavior. \newline
% -> In summary, we might have a preference for the way we modify the control signals of a nominal controller in the context of pHRI. \newline
% -> explicitly mention the main contributions of this work: 1) we propose, for the first time, using hierarchical qp to solve CBFs of multiple tasks of different priority levels for safe pHRI. 2) we used extensive simulations and/or experiments to demonstrate the effectiveness of the proposed method, and also showed the possibility of handling soft safety tasks whose safety limits did not have to be explicitly specified in advance, i.e., time-varying CBFs.
% }

In the era of ubiquitous robots, various types of intelligent robots are being integrated into our work and daily lives. Physical human-robot interaction (pHRI) is becoming an important form of human-robot coexistence for performing challenging tasks (e.g., human-robot co-manipulation), assist/augment human performance (e.g., robotic exoskeleton), and study human sensorimotor control (e.g., human impedance regulation) \cite{fang2023human}. Safety is of paramount importance for the deployment of pHRI.

Control barrier functions (CBFs) \cite{ames2019control} represent a very successful technique in safety-critical applications as, unlike conservative approaches such as potential fields \cite{Fang2010new} or offline constraint planning, they i) ensure continuous-time forward invariance of the safety set;
%(similarly to what control Lyapunov functions (CLFs) do for stability); 
and ii) can be integrated with any task-based nominal controller, possibly designed independently from the CBF, along a quadratic optimization program which minimally modifies the input at real-time frequencies. CBFs have been widely used to enforce kinematics-related safety constraints, e.g., collision avoidance with the environment, including obstacles \cite{singletary2021safety, sun2023adaptive, morton2025safe} and humans \cite{ferraguti2020safety, maithani2026proactive}. Beyond collision avoidance, control barrier functions have also been successfully applied to dynamics-related safety constraints, e.g., robot energy limiting \cite{michel2024novel, califano2025limiting}. However, no approach has encoded safety measures directly into CBFs for continuous pHRI where the interaction between the human and the robot is actively utilized rather than being avoided, e.g., in rehabilitation applications.

% safety-critical problems such as physical human–robot interaction. While in some isolated attempts CBFs have recently been formulated to encode direct energy and power constraints (A Novel Safety‑Aware Energy Tank Formulation Based on Control Barrier Functions (Michel, Saveriano  and Lee), Limiting Kinetic Energy Through Control Barrier Functions (Califano, Logmans and Roozing)) such energy‑based safety objectives have not been integrated into physical human–robot interaction scenarios.

In most prior work, all CBF safety constraints are formulated as a set of inequality constraints, typically solved together in a single QP, which is desirable when all of these constraints are equally important and strict. However, when we apply the CBF framework to the safe continuous pHRI applications with one QP for multiple different safety tasks, we might have some interaction tasks that we want to preserve as much as possible, and we prioritize the modification of the null-space joint velocity/torque component over that of the joint velocity/torque component that would affect the desired interaction behavior.
In short, we may have a preference for how we modify the control signals of a nominal controller minimally in the context of pHRI. 

Driven by this motivation, the main contributions of this work include: 

1) We proposed, for the first time, using hierarchical QP to solve CBFs of multiple safety and performance tasks of different priority levels for safe continuous pHRI; 

2) We used comparative experiments to demonstrate the effectiveness of the proposed method against a baseline method where all the safety tasks are treated equally.

\begin{figure}[!t]
    \centering
    \includegraphics[trim = 20mm 0mm 0mm 0mm, clip, scale = 0.29]{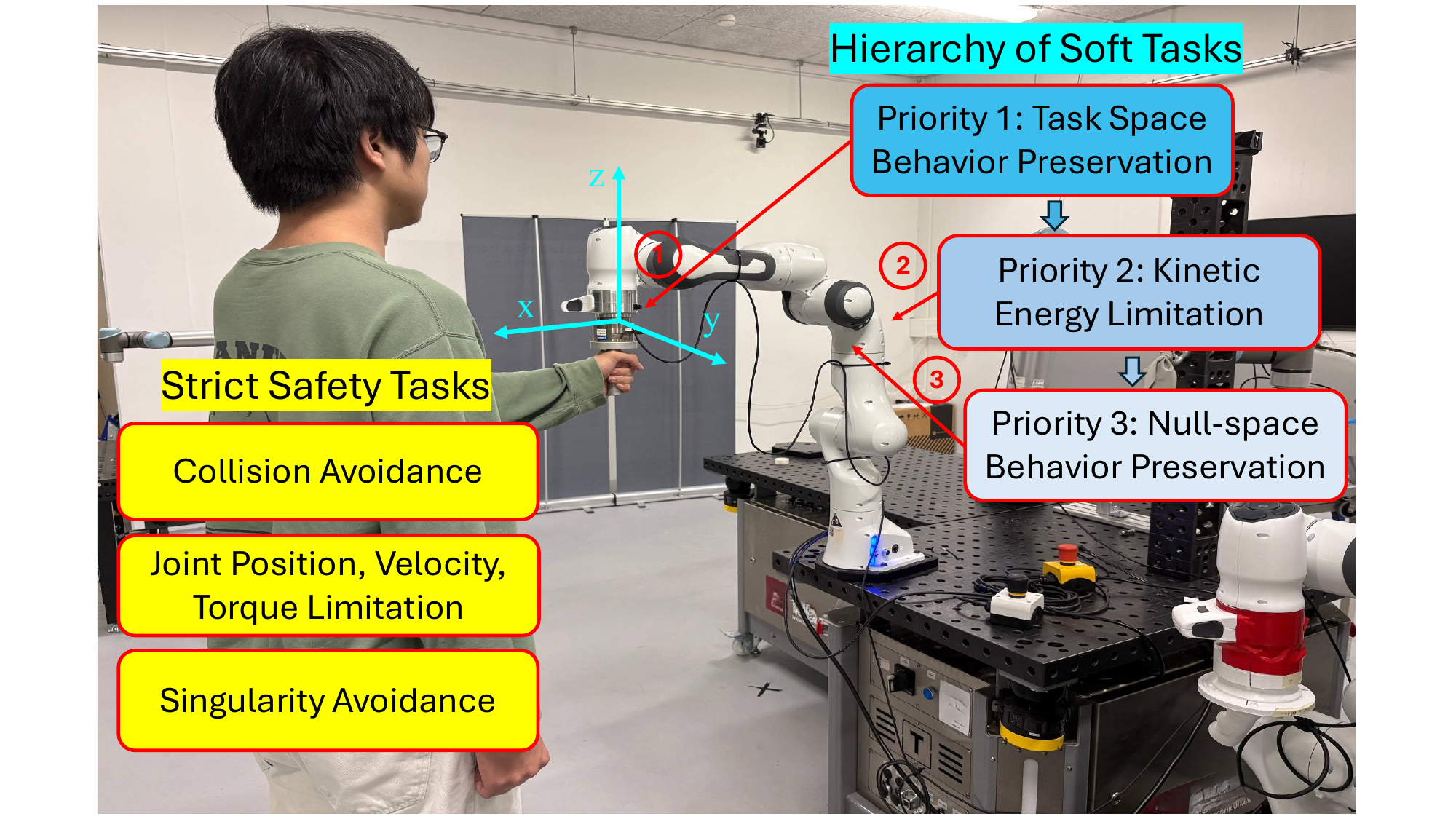}
    \caption{A real physical human-robot interaction scenario with multiple strict safety tasks and a hierarchy of soft tasks of different priority levels.}
    \label{fig:Soft tasks and strict tasks}
\end{figure}

\section{Related works}

% \textcolor{blue}{a suggested structure:\newline
% -> (first paragraph) SoTA on how people used CBFs for safety-critical applications.\newline
% -> (second paragraph) SoTA on how people used hierarchical qp for the motion control of highly redundant robots, e.g., humanoid robots. \newline
% -> (third paragraph) did anyone solve CBFs with hierarchical qp before? if nobody did that, we stress the motivation again for safe pHRI with a smooth transition to Section III. 
% }

Control barrier functions offer formal guarantees of forward invariance of safe sets and are well-suited for real‑time implementation. Originally developed in a control-theoretic setting to enforce safety constraints via forward‑invariance conditions, CBFs gained popularity through a CBF‑CLF‑QP framework \cite{taylor2020adaptive}, where safety encoded by CBF inequalities is combined with stability objectives encoded by control Lyapunov functions (CLFs) in a single quadratic program (QP) to be solved in real-time. In robotics, CBFs have been applied across a range of applications, from autonomous vehicle safety (e.g., cooperative adaptive cruise control) \cite{ames2014control} to legged locomotion \cite{kim2024safety}. Across these works, multiple safety constraints encoded by different CBFs are normally strictly respected and treated equally important in a single QP. As a consequence, no prioritization structure for multiple safety constraints was considered. 
%A constraint can be relaxed by introducing a slack variable, a common trick to increase the feasible solution set in QPs. 

When a robot has redundant Degrees of Freedom (DoFs), it can simultaneously perform multiple tasks, including not only the most important safety-related tasks but also other less strict tasks that can tolerate errors, for instance, trajectory-tracking tasks. To ensure that the execution of less important tasks does not affect the performance of more important tasks, hierarchical QP has been widely employed to manage and solve a hierarchy of tasks of different priorities. A hierarchy of equality constraints was designed and solved recursively using nullspace projectors to ensure that a solution to a lower-priority task can only be found in the nullspace of a higher-priority task \cite{slotine1991general}. The computational efficiency can be improved by using the basis vectors of the nullspace rather than the projectors in the recursive calculations \cite{escande2010fast}. The equality hierarchical QP was extended further to incorporate a system of inequality constraints, becoming a more general and flexible framework in \cite{mansard2009unified, kanoun2011kinematic, fang2015efficient} with similar efforts improving computation efficiency for real-time applications of a robot with a larger number of DoFs, such as a humanoid robot \cite{de2010feature, mansard2012dedicated, escande2014hierarchical, zhou2016generic}. The QP-based hierarchy was primarily used for kinematic control of redundant robots, and its major advantages are that inequality constraints can be directly incorporated into the optimization problem, facilitating the incorporation of hardware limits, e.g., joint position/velocity limits and singularity avoidance. Nevertheless, one has to compute the complete inverse dynamics to drive the actuators to track the generated motion. In addition, external interaction forces are not considered, and a formal stability proof for the closed-loop system is not provided.

With the desirable and embedded stability of a dynamic system and compatibility with a real-time QP framework in the CBF formulation, a hierarchical QP can be used to solve multiple CBF-formulated safety tasks of different priorities. To the best of the authors' knowledge, very few researchers tried to use the hierarchical strategy to solve multiple CBF-formulated tasks. In \cite{lee2023hierarchical, maithani2026proactive}, a concept of hierarchical CBF was proposed to allow hierarchical relaxation of a group of safety tasks to enforce a hierarchy of these tasks. However, only safety tasks were included in the hierarchy, and also only kinematics-based CBFs were considered in their applications, i.e., navigation of a legged dog in a cluttered environment and collision avoidance between a moving human body and a robot arm. In this work, we aim to design a general and flexible framework that uses hierarchical QP to solve kinematics- and energy-based CBF-formulated safety tasks and performance tasks of different priority levels for safe pHRI, where physical interaction is continuously maintained.  

\section{Problem Formulation}

\subsection{Control barrier function-quadratic programming framework}

We consider a robotic system with a generalized state
$x \in \mathbb{R}^n$ and control input $u \in \mathbb{R}^n$.
To impose a safety-related constraint, Control Barrier Function (CBF) provides a systematic approach to enforce forward invariance
of a prescribed safe set.

\subsubsection{Control Barrier Function}

Let a CBF, $h(x)$, be a continuously differentiable function defining a safe set,
\[
\mathcal{C} = \{ x \mid h(x) \ge 0 \}.
\]
Forward invariance of this safe set can be guaranteed by imposing an inequality constraint,
\begin{equation}
\dot h(x,u) + \gamma h(x) \ge 0,
\end{equation}
where $\gamma > 0$ is a class-$\mathcal{K}$ function.
For control-affine systems, this condition can be rewritten as an affine inequality of the control input,
\begin{equation}
A_h(x) u \ge b_h(x).
\end{equation}

The CBF is then solved in a Quadratic Programming (QP) problem formulated below for a safe control input $u$,
\begin{equation}
\label{eq:qp}
\min_u \; \| u - u_{\mathrm{nom}} \|^2
\quad
\text{s.t.}
\quad
A_h(x) u \ge b_h(x),
\end{equation}
to obey the CBF-formulated safety inequality constraint while being as close to the control command of a nominal controller, $u_{\mathrm{nom}}$, as possible.

\subsubsection{Perspective on the quadratic programming problem}

The objective function of the QP problem in (\ref{eq:qp}), $\min \|u - u_{\mathrm{nom}}\|^2$, is often interpreted as a soft equality task. In a general form, this type of task can be formulated as: 
\begin{equation}
\label{eq:softe}
\min_u \; \| A_e(x) u - b_e(x) \|^2,
\end{equation}
where $A_e(x)$ is an identity matrix and $b_e(x)$ is $u_{\mathrm{nom}}$ in (\ref{eq:qp}), meaning that an error can be tolerated, but is minimized as much as possible. Alternatively, equality tasks can also be enforced strictly as a constraint of a standard QP problem, i.e., $A_e(x)u=b_e(x)$. Strict equality tasks are needed, for instance, when accurate velocity tracking of a robot arm's end-effector is required via a Jacobian that maps joint velocities to the desired end-effector velocity. Hence, an equality task can appear in a QP problem either as the objective function for a soft task or as a constraint for a strict task.

Similarly, the constraint component of the QP problem in (\ref{eq:qp}),
can be interpreted as an inequality task, which defines an admissible set of the optimization variable:
\begin{equation}
A_i(x) u \ge b_i(x).
\end{equation}
When enforced strictly, the constraint defines a safety boundary, i.e., keeping the system state within a safe set. This is typically used when implementing safety requirements that must never be violated, such as collision avoidance. However, in some cases, to maintain the feasibility of a solution with multiple conflicting constraints, the inequality task may be relaxed through a positive slack variable $\delta$,
\begin{equation}
\label{eq:qp2}
\min_{u,\delta} \; \| \delta \|^2
\quad
\text{s.t.}
\quad
A_i(x) u \ge b_i(x) - \delta,
\end{equation}
with $\delta$ penalized in the objective function. Such soft inequality tasks are suitable for constraints that can tolerate limited violation,
for instance, kinetic energy or power regulation, whose exact upper limits are difficult to define. Therefore, an inequality task can appear in a QP problem either as a constraint for a strict task or as the objective function for a soft task.

\subsection{Task Categorization}\label{sec:taskcat}

Because of the convertibility between the objective function and the constraint in a QP problem for describing a task mentioned above, we will use \textit{Task} as a unified term instead of the objective function and constraint in the following text to avoid ambiguity. In this work, tasks in a CBF-QP framework are categorized from three different perspectives:

\subsubsection{Equality vs. Inequality tasks}

A task can be formulated as an equality or inequality with respect to the control input, $u$. For example, an equality task may arise directly from the controller design, such as tracking a desired Cartesian-space velocity or force, with the joint velocity or torque as the control input. An inequality task may be derived from the CBF, or it can come from hardware limits, such as joint velocity and/or torque limits.

\subsubsection{Strict vs. Soft tasks}

A task can be strict or soft, depending on whether an error in the task is tolerated. A strict task is more important than a soft task and must be respected at all times, for example, to satisfy hardware limits and avoid collisions. In a soft task, the error must be minimized as much as possible.

\subsubsection{Safety vs. Performance tasks}

A safety task can be formulated directly from a CBF, for instance, an energy-limiting task, and is also related to hardware limits. In contrast, a performance task is associated with the behavior of the nominal controller, i.e., the final control signal is preferred to be as close as possible to the nominal control signal. 

Based on the aforementioned definitions, the objective function of (\ref{eq:qp}) in a typical CBF-QP framework is a soft, equality, performance task, while the constraint of (\ref{eq:qp}) is a strict, inequality, safety task. This flexible task-categorization method provides a unified view of task treatment within the CBF-QP optimization framework.

% Motivated by the CBF-QP structure, optimization tasks in robotic control
% can be categorized as either equality or inequality tasks with respect to the control input. Furthermore, tasks can also be classified as strict or soft, depending on whether violation is permitted.

% This flexible treatment of equality and inequality tasks provides a unified view of optimization problems in robotic control. Within a hierarchical QP framework, tasks can be systematically organized according to their priority and strictness, allowing them to transition between objective terms and constraints as required. This perspective naturally motivates the hierarchical formulation introduced in the sequel.

\subsection{Motivation for a hierarchy of multiple tasks}

Although all safety and performance tasks can be formulated in a single QP problem, solving them within a one-layer optimization framework is often inadequate. Such a formulation suffers from several inherent issues.

\subsubsection{Issue 1: Deviation from desired task-space behavior}

When humans physically interact with a redundant robot, we may want to preserve the dynamic behavior of a nominal controller at the human-robot contact point as much as possible, and utilize the nullspace of the contact point to satisfy all safety tasks. Minimizing $\|u - u_{\text{nom}}\|^2$ in joint space does not guarantee preservation of the behavior, as small deviations in joint space may induce significant distortions in the dynamic behavior at the contact point.

To preserve the performance of a nominal controller at the contact point, e.g., a Cartesian impedance controller, one can decompose the control input into task-space and nullspace components using the dynamically consistent projection matrix $P$ \cite{dietrich2015overview}. The control input can be split as:
\begin{equation}
u = P u + N u,
\end{equation}
where $P$ projects onto the task space of the contact point, and $N = I - P$ projects onto the nullspace of the contact point. To minimize the deviation from the desired task-space behavior, one can design a soft, equality, performance task $\|Pu - Pu_{\text{nom}}\|^2$, and prioritize modifying the nullspace component, $N u$, to satisfy other strict tasks. 

%------------------------------------------------
\subsubsection{Issue 2: Infeasibility and uncertain limits of safety metrics}

In practice, multiple safety tasks may conflict, resulting in an empty feasible set of control inputs. Furthermore, the limits of certain safety metrics may not be precisely known. For example, A reasonable kinetic energy limit may vary depending on the interaction conditions and the injury biomechanics of the individuals involved.

To address infeasibility and uncertain boundaries, strict tasks can be relaxed and transformed into soft tasks using (\ref{eq:softe}) and (\ref{eq:qp2}). The soft task error is minimized at each control cycle, allowing temporary relaxation only when strictly necessary. This ensures feasibility while keeping boundaries of the safety metrics as tight as possible.

% \begin{equation}
% A_i(x) u \ge b_i(x),
% \end{equation}

% the condition is relaxed as

% \begin{equation}
% A_i(x) u \ge b_i(x) - \delta,
% \quad \delta \ge 0.
% \end{equation}

%------------------------------------------------
\subsubsection{Issue 3: Lack of explicit task priority}

Given the need for multiple performance and soft tasks, solving them within a single optimization problem with different weights can be problematic, as their relative importance cannot be enforced strictly. Critical safety tasks may be unintentionally compromised to satisfy secondary tasks. For example, collision avoidance must be prioritized over energy regulation, since physical contact may cause immediate damage, whereas a temporary increase in energy may be tolerable. 

To ensure a clear priority structure and facilitate the management of a large number of tasks, all tasks can be organized into a hierarchy of different priority levels. Higher-priority tasks are enforced first, and lower-priority tasks are optimized only within the feasible subset defined by higher-priority tasks. This strict hierarchical structure guarantees that higher-priority tasks are never sacrificed for lower-priority tasks. These needs motivate us to adopt the \textit{Hierarchical Quadratic Programming (HQP)} framework \cite{kanoun2011kinematic}, which provides a rigorous mechanism to enforce hierarchical priority while preserving solution feasibility in the presence of conflicting tasks with uncertain limits.

%------------------------------------------------
% By integrating task-space decomposition, strict priority ordering,
% and adaptive slack relaxation, the multi-objective control problem
% is naturally formulated within a hierarchical optimization structure.
% Instead of aggregating all objectives into a single weighted program,
% the control objectives are arranged into multiple priority levels,
% which are solved sequentially.
% At each level, the feasible set is restricted by the optimal solution
% and constraints of all higher-priority levels.

\section{Hierarchical quadratic programming}

\subsection{General framework}

The section presents a concise overview of a general framework for the Hierarchical Quadratic Programming, originally developed by Kanoun O. et al. for kinematic control of a humanoid robot in \cite{kanoun2011kinematic}. HQP is an optimization framework that enforces a prioritized structure among multiple kinematic tasks. Rather than combining all tasks into a single quadratic program, HQP decomposes the problem into a sequence of quadratic programs organized by different priority levels. Each level is solved in order, from highest to lowest priority. The admissible set of the optimizable control input after a lower-priority task has been solved would always be a subset of the admissible set after all the higher-priority tasks have been solved. In this way, a strict hierarchy of multiple tasks can be ensured.

The initial admissible set $S_0$ of the control input is defined by strict equality and inequality tasks:
\begin{equation}
S_0 =
\left\{
u \;\middle|\;
A_0 u = b_0,\;
C_0 u \ge d_0
\right\},
\end{equation}
which characterizes the system's initial admissible region, which is a convex polytope. These strict tasks encode physical limitations, actuation bounds, and other non-negotiable conditions. They must be satisfied at all priority levels and cannot be relaxed.

\subsubsection{Optimization at priority level $k$.}

Given the feasible set $S_{k-1}$, the control input at level $k$ is optimized by solving
\begin{equation}
\begin{gathered}
(u_k^*, \delta_k^*)
= \arg\min_{(u\in S_{k-1}, \delta_k)}
\frac{1}{2}\|A^{\mathrm{new}}_k u - b^{\mathrm{new}}_k\|^2 + \frac{\rho_k}{2}\delta_k^2, \\
\text{s.t.} ~~~~ C^{\mathrm{new}}_ku \ge d^{\mathrm{new}}_k-\delta_k.
\end{gathered}
\end{equation}
where a soft equality task and a soft inequality task are optimized together with $\rho_k > 0$ penalizing the relaxation of the inequality task. The slack variable $\delta_k$ allows controlled relaxation of the soft inequality tasks. The relative importance between inequality and equality tasks can be adjusted via $\rho_k$, thereby balancing the tasks according to design requirements.

\subsubsection{Recursive construction of admissible sets.}

Assume that the admissible set at level $k-1$ is represented as
\begin{equation}\label{eq:sk-1}
S_{k-1}=
\left\{
u \;\middle|\;
A_{k-1}u=b_{k-1},\;
C_{k-1}u\ge d_{k-1}
\right\}.
\end{equation}
After solving the QP problem at level $k$, to update the admissible set, two constraints are added: an equality constraint that inherits from the equality task at level $k$ and an inequality constraint that inherits from the inequality task at level $k$, thereby preserving higher-priority task solutions:
\begin{equation}
\begin{aligned}
A_k^{\mathrm{new}}u &=A_k^{\mathrm{new}}u^*, \\
C_k^{\mathrm{new}}u &\ge d_k^{\mathrm{new}}-\delta^*_k.
\end{aligned}
\end{equation}

Then the constraint matrices in (\ref{eq:sk-1}) at level $k$ are updated by stacking:
\begin{equation}
A_k=
\begin{bmatrix}
A_{k-1}\\
A_k^{\mathrm{new}}
\end{bmatrix},
\qquad
b_k=
\begin{bmatrix}
b_{k-1}\\
A_k^{\mathrm{new}}u^*
\end{bmatrix},
\label{eq:Ak_update}
\end{equation}
\begin{equation}
C_k=
\begin{bmatrix}
C_{k-1}\\
C_k^{\mathrm{new}}
\end{bmatrix},
\qquad
d_k=
\begin{bmatrix}
d_{k-1}\\
d_k^{\mathrm{new}}-\delta^*_k
\end{bmatrix}.
\label{eq:Ck_update}
\end{equation}

The admissible set after solving the $k$-th QP is updated accordingly:
\begin{equation}
S_k=
\left\{
u\;\middle|\;
A_ku=b_k,\;
C_ku\ge d_k
\right\}.
\end{equation}

By construction, these constraint matrices are monotonically augmented as a sequence of QP problems are solved, and the admissible set gradually shrinks, i.e., $S_k\subseteq S_{k-1}$. As a result, lower-priority tasks are optimized only within the admissible set defined by higher-priority levels, thereby ensuring the intended hierarchical ordering of tasks. In this framework, note that all strict tasks should go to $S_0$, while all soft tasks are managed in the rest of the hierarchy at different priority levels.  

\subsection{Application of the CBF-HQP in pHRI}

We propose using the HQP framework as a unified yet flexible hierarchical structure to manage multiple tasks across various categories (defined in Section \ref{sec:taskcat}) in pHRI, including not only kinematics-based but also energy-based CBF-formulated safety tasks. In addition, one distinct feature of this work compared to the previous similar works based on hierarchical CBF \cite{lee2023hierarchical, maithani2026proactive} is that our proposed method allows both safety and performance tasks to be designed at any level of the hierarchy in a flexible way to balance the safety and performance based on different applications.  

In this section, a specific pHRI scenario is used as an example to illustrate the application of the HQP framework. We consider a scenario in which a human operator physically interacts with a redundant robot arm via a handle mounted on the robot's end-effector, maintaining continuous contact (see Fig. \ref{fig:Soft tasks and strict tasks}). This pHRI scenario could be relevant for rehabilitation applications, for instance.

\subsubsection{Relevant tasks}

The relevant tasks in this scenario are listed as follows:

\textit{Strict safety tasks}: The joint position, velocity, torque limits and collision avoidance must never be violated. These tasks define a hard safety boundary.

\textit{Soft inequality safety tasks}: CBF-formulated kinetic energy limiting task. Since the kinetic energy bound may be uncertain or time-varying, this task can be formulated as a soft task, allowing relaxation via a slack variable when necessary to preserve feasibility.

\textit{Soft equality performance tasks}: The Cartesian space and nullspace performance tasks, which aim to preserve the desired interactive behavior of a nominal controller, e.g., a Cartesian impedance controller. 

%\vspace{0.5cm}
\subsubsection{Priority hierarchy design}

To ensure safety, all the strict safety tasks must always be satisfied. Apart from that, in an application where the performance task is prioritized, the objective is to minimize modifications to the component of the control input that affects the dynamic behavior at the human–robot interaction point, while minimizing the overall kinetic energy. To achieve this, a hierarchical priority structure can be designed as follows:

\textit{Initial admissible set ($S_0$)}: All the strict safety tasks will be used to form $S_0$, while all the soft tasks will be designed at priority levels of the HQP.

\textit{Priority 1 (Cartesian-space behavior preservation)}: Since the human-robot interaction point is at the handle, and the desired dynamic behavior of a nominal controller can be preserved as much as possible by (joint torque is assumed to be used as the control input $u$):
\[
\min \; \|P u - P u_{\mathrm{nom}}\|^2.
\]
%so that the desired interaction behavior is preserved.

\textit{Priority 2 (Kinetic-energy limiting)}: Whole-body kinetic energy is limited through a CBF-formulated soft inequality safe task. The energy limit can be relaxed minimally to preserve feasibility under uncertainty when necessary.

\textit{Priority 3 (Nullspace behavior preservation)}: The nullspace behavior preservation is set as the lowest-priority task by minimizing
\[
\min \; \|N u - N u_{\mathrm{nom}}\|^2.
\]

\subsubsection {Formulation of cascaded QPs}
\begin{figure}[htbp]
    \centering
    \includegraphics[trim = 15mm 0mm 18mm 0mm, clip, scale = 0.27]{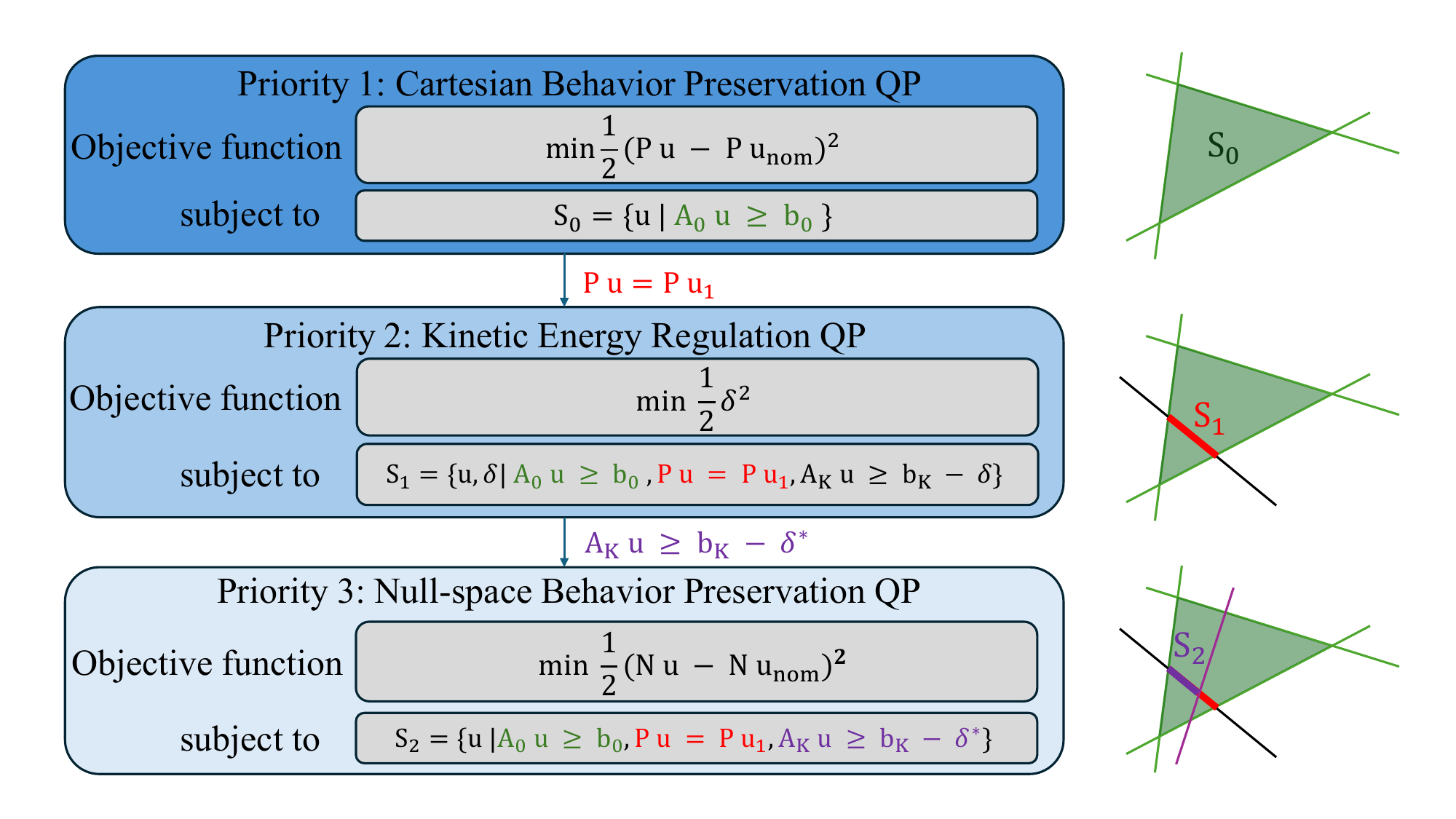}
    \caption{Diagram of a cascaded structure of the CBF-HQP for a physical human-robot interaction scenario.}
    \label{fig:Structure of HQP for a specific robot}
\end{figure}

Based on the relevant task and the designed priority structure above, we now formulate the QP problem at each level of the HQP. An overview of the formulation is illustrated in Fig.~\ref{fig:Structure of HQP for a specific robot},
which depicts the layered structure and the propagation of admissible sets.

Before constructing the hierarchical layers, we first define the initial admissible set $S_0$. All strict safety tasks, including collision avoidance, joint velocity  torque limits, are expressed in a unified affine inequality form as
\begin{align}
A_0(x)\,u \ge b_0(x),
\end{align}
where $A_0(x)$ and $b_0(x)$ are obtained by stacking the corresponding inequality rows. In particular, collision avoidance is enforced by a CBF, while joint torque limits represent actuator constraints. These constraints define the initial admissible set,
\[
S_0 = \{\, u \mid A_0(x)u \ge b_0(x) \,\},
\]
which is enforced at all priority levels without relaxation.

\textit{Priority 1: Cartesian-space behavior preservation}

At the highest priority level, we minimize the deviation of the Cartesian-space component within the initial admissible set $S_0$:
\begin{equation}
u_1
=
\arg\min_{u \in S_0}
\| P u - P u_{\mathrm{nom}} \|^2.
\end{equation}

\textit{Priority 2: Kinetic energy regulation}

To regulate the whole-body kinetic energy while preserving feasibility under uncertain safety bounds, we introduce a slack variable $\delta \ge 0$ and define the optimization variable

\begin{equation}
z = \begin{bmatrix} \delta \\ u \end{bmatrix}.
\end{equation}

The total kinetic energy of the manipulator is defined as
\begin{equation}
K(q,\dot q) = \frac{1}{2}\dot q^\top M(q)\dot q .
\end{equation}

Taking its time derivative gives
\begin{equation}
\dot K
=
\dot q^\top M(q)\ddot q
+
\frac{1}{2}\dot q^\top \dot M(q)\dot q .
\end{equation}
Using the standard property that $\dot M(q)-2C(q,\dot q)$ is skew-symmetric,
one has
\begin{equation}
\frac{1}{2}\dot q^\top \dot M(q)\dot q
=
\dot q^\top C(q,\dot q)\dot q ,
\end{equation}
and therefore
\begin{equation}
\dot K
=
\dot q^\top \big(M(q)\ddot q + C(q,\dot q)\dot q\big).
\end{equation}

Substituting the robot dynamics
\begin{equation}
M(q)\ddot q + C(q,\dot q)\dot q + g(q)
= \tau + \tau_{\mathrm{ext}},
\end{equation}
yields
\begin{equation}
\dot K
=
\dot q^\top \big(\tau + \tau_{\mathrm{ext}} - g(q)\big).
\end{equation}

To regulate the kinetic energy with minimal relaxation,
a slack variable $\delta \ge 0$ is introduced and the relaxed CBF is defined as
\begin{equation}
h_K = K_{\max} - K + \delta .
\end{equation}

Applying the exponential CBF condition
\begin{equation}
\dot h_K + \gamma h_K \ge 0,
\end{equation}
together with the discrete-time approximation
\begin{equation}
\dot \delta \approx \frac{\delta - \delta^-}{\Delta t},
\end{equation}
leads to
\begin{equation}
\left(\gamma + \frac{1}{\Delta t}\right)\delta
- \dot q^\top \tau
\ge
-\gamma(K_{\max} - K)
+ \frac{\delta^-}{\Delta t}
+ \dot q^\top (\tau_{\mathrm{ext}} - g(q)).
\end{equation}

where $\delta^{-}$ denotes the slack value from the previous time step.The above inequality can be compactly written in affine form as

\begin{equation}
A_K u + c_K\,\delta \ge b_K.
\end{equation}

%The above inequality with the solution of  is incorporated into the admissible set of Layer~2.
To preserve the higher-priority Cartesian-space behavior obtained in Priority~1,
We enforce the equality task inherited from Priority 1,
\begin{equation}
P u = P u_1.
\end{equation}
The admissible set at Priority~2 is thus defined as
\begin{equation}
S_1 =
\left\{
(u
\;\middle|\;
\begin{aligned}
& u \in S_0, \\
& P u = P u_1. \\
\end{aligned}
\right\}.
\end{equation}

At Priority~2, we minimize the slack variable in order to reduce the relaxation of the kinetic-energy constraint:
\begin{equation}
\begin{aligned}
(u_2, \delta^*)
&=
\arg\min_{(u\in S_1, \delta)}
\frac{1}{2}\delta^2, \\
\text{s.t.} ~~  A_K u + c_K\,\delta &\ge b_K, \\
 \delta &\ge 0
\end{aligned}
\end{equation}

\textit{Priority 3: NullSpace behavior preservation}

Finally, we utilize the remaining redundancy to preserve the nullspace behavior of the nominal controller as much as possible while respecting all higher-priority tasks. The admissible set at this level is constructed by inheriting from the solution of the QP at Priority 2 with the optimal slack value $\delta^*$ set as a constant:
\begin{equation}
S_2 =
\left\{
u
\;\middle|\;
\begin{aligned}
& u \in S_0, \\
& P u = P u_1, \\
& A_K u \ge b_K - c_K \delta^*
\end{aligned}
\right\}.
\end{equation}

The nullspace optimization is then formulated as

\begin{equation}
u_3 =
\arg\min_{u \in S_2}
\| N u - N u_{\mathrm{nom}} \|^2.
\end{equation}

This guarantees that the nullspace behavior is optimized only within the admissible region defined by all the higher-priority tasks. The resulting control input $u_3$ is then applied to the robot as the final command.

% \subsection{Efficiency improvement}
% \href{red}{\textit{Introduce a more efficient way to do hierarchical QP(Do the 6D first, and do the null-sapce 4D in the next two layers)
% }}

\section{Experimental Validation}

To evaluate the effectiveness of the proposed hierarchical CBF framework, experiments were conducted on a torque-controlled robot manipulator. Two experiments were designed to validate the proposed framework:

1) Experiment 1 - Step change in equilibrium position:
The desired equilibrium position of the Cartesian impedance controller of the robot
was abruptly changed to a new constant position.

2) Experiment 2 - Sinusoidal external force:
For reproducibility, a virtual sinusoidal external force was applied at the end-effector to simulate continuous pHRI.

In both cases, the proposed hierarchical QP controller was compared with a single-layer QP formulation, where all tasks are optimized at the same level. Experiment~2 further demonstrated the flexibility of the HQP framework, as the task hierarchy can be reconfigured,
e.g., assigning the kinetic-energy safety constraint to Priority~1 and the Cartesian behavior preservation to Priority~2.

\subsection{Experimental Setup}

Experiments were performed on a Franka Emika Panda 7-DoF robotic manipulator operating in torque control mode. The Franka Control Interface (FCI) provides a torque control interface at  $1\,\mathrm{kHz}$, with built-in gravity and friction compensation enabled by default. 
Joint positions $q$, joint velocities $\dot{q}$, and external torque estimates 
$\hat{\tau}_{\mathrm{ext}}$ are available through the interface. The nominal control input was generated by a Cartesian impedance controller with a translational stiffness of $200\,\mathrm{N/m}$ and a rotational stiffness of $50\,\mathrm{Nm/rad}$ in all directions,
with critically damped dynamics.

\subsection{Experiment 1: Step change in equilibrium position}

In this experiment, the equilibrium position of the Cartesian impedance controller was abruptly displaced by 0.2m in the z direction, inducing a virtual spring force that drove the robot towards the new equilibrium. This sudden change injected elastic potential energy into the system, which was rapidly converted into kinetic energy, potentially leading to large transient motions.

\begin{figure}[!t]
    \centering
    \includegraphics[width=\columnwidth]{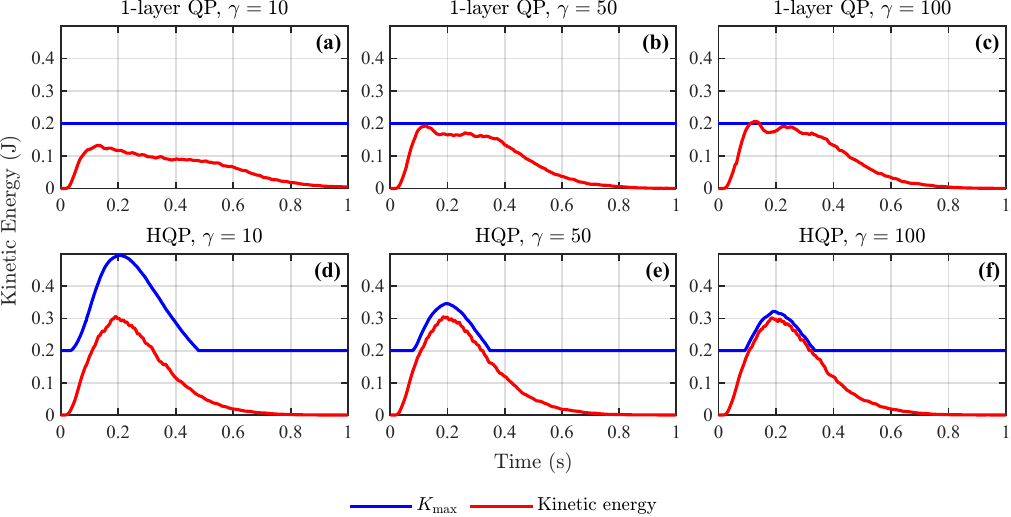}
    \caption{Experiment 1 (Equilibrium shift): total kinetic energy.}
    \label{fig:Experiment1_1}
\end{figure}

\begin{figure}[!t]
    \centering
    \includegraphics[width=\columnwidth]{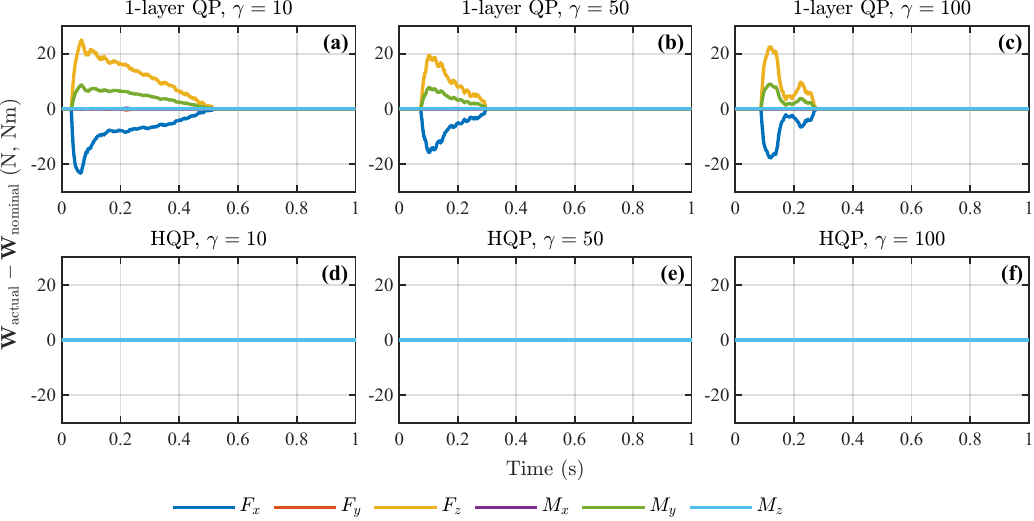}
    \caption{Experiment 1 (Equilibrium shift): Cartesian wrench deviation.}
    \label{fig:Experiment1_2}
\end{figure}

\begin{figure}[!t]
    \centering
    \includegraphics[width=\columnwidth]{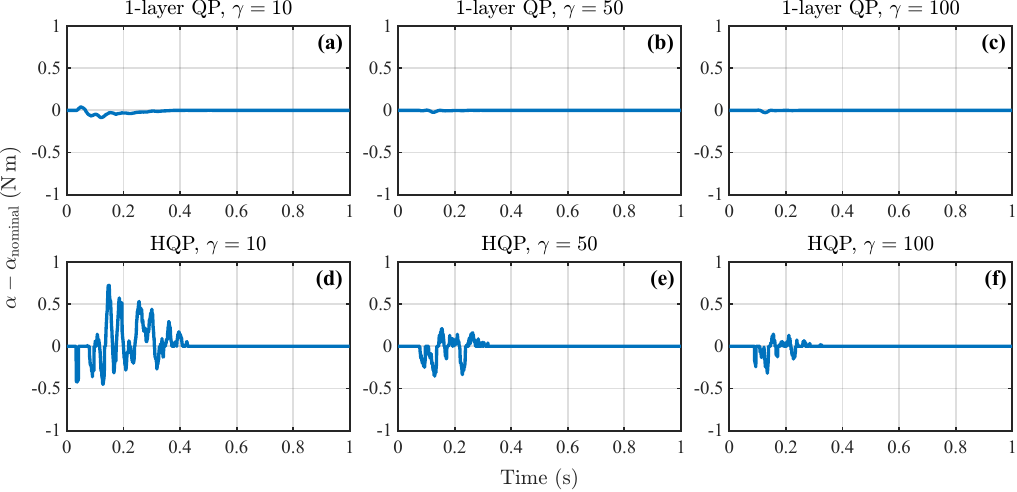}
    \caption{Experiment 1 (Equilibrium shift): Null-space component.}
    \label{fig:Experiment1_3}
\end{figure}

Figure~\ref{fig:Experiment1_1} compares the single-layer QP formulation and the proposed hierarchical QP under three different values of CBF gain $\gamma$.
In the single-layer formulation, the kinetic energy was directly constrained via the CBF condition; therefore, the kinetic energy remained strictly within the predefined bound throughout the motion (Fig.~\ref{fig:Experiment1_1}(a--c)).
However, this strict enforcement can significantly modify the nominal Cartesian impedance behavior. This effect arised from the structure of the optimization problem, where the control input was obtained by solving
\[
\min_{u} \| u - u_{\mathrm{nominal}} \|^2
\]
subject to the CBF-based kinetic energy constraint. A minimal modification to the joint-space control input does not ensure a minimal modification to the Cartesian-space dynamics.

In contrast, within the proposed HQP structure, kinetic energy regulation was given lower priority than preserving Cartesian behavior. Moreover, we introduced a slack variable to handle the kinetic-energy upper bound, which can be hard to determine. As shown in the second row of Fig.~\ref{fig:Experiment1_1}(d--f), the maximum kinetic energy $K_{\max}$ can be adjusted dynamically through the slack variable.
This design minimizes the excess while avoiding overly conservative behavior.

Figure~\ref{fig:Experiment1_2} illustrates the difference between the optimized Cartesian equivalent wrench and the nominal Cartesian equivalent wrench,
i.e., $\Delta \mathbf{W} = \mathbf{W}_{\mathrm{actual}} - \mathbf{W}_{\mathrm{nominal}}$.
In the HQP framework, Cartesian behavior preservation was treated as the highest-priority objective; consequently, the wrench deviation remained identically zero, indicating that the nominal Cartesian behavior was preserved. In contrast, the single-layer CBF-based optimization substantially altered the Cartesian wrench to satisfy the kinetic energy bound, leading to a non-zero wrench deviation.

Figure~\ref{fig:Experiment1_3} illustrates the null-space behavior.
The control torque is decomposed as
\[
\tau = P u_{\mathrm{nom}} + \alpha z,
\]
where $z$ denotes a null-space basis vector and $\alpha$ represents the corresponding null-space coefficient.

The deviation $\alpha - \alpha_{\mathrm{nom}}$ was plotted in Figure~\ref{fig:Experiment1_3} to evaluate how the null-space component was modified under different optimization strategies. In the single-layer formulation,
The deviation remained approximately zero, indicating that the null-space objective was largely preserved. In contrast, with the proposed HQP framework, $\alpha$ exhibited significant variations during the step excitation.
This behavior indicated that the controller sacrificed nominal null-space behavior to strictly preserve higher-priority tasks and to regulate the kinetic energy, which was consistent with the intended hierarchical design.

Overall, these results demonstrated that the proposed HQP could achieve strict task prioritization and support flexible handling of strict and soft tasks: high-priority Cartesian behavior could be preserved, while the kinetic energy was regulated with adaptive flexibility via slack variables.

\subsection{Experiment 2: Sinusoidal external force}

\begin{figure}[!t]
    \centering
    \includegraphics[width=\columnwidth]{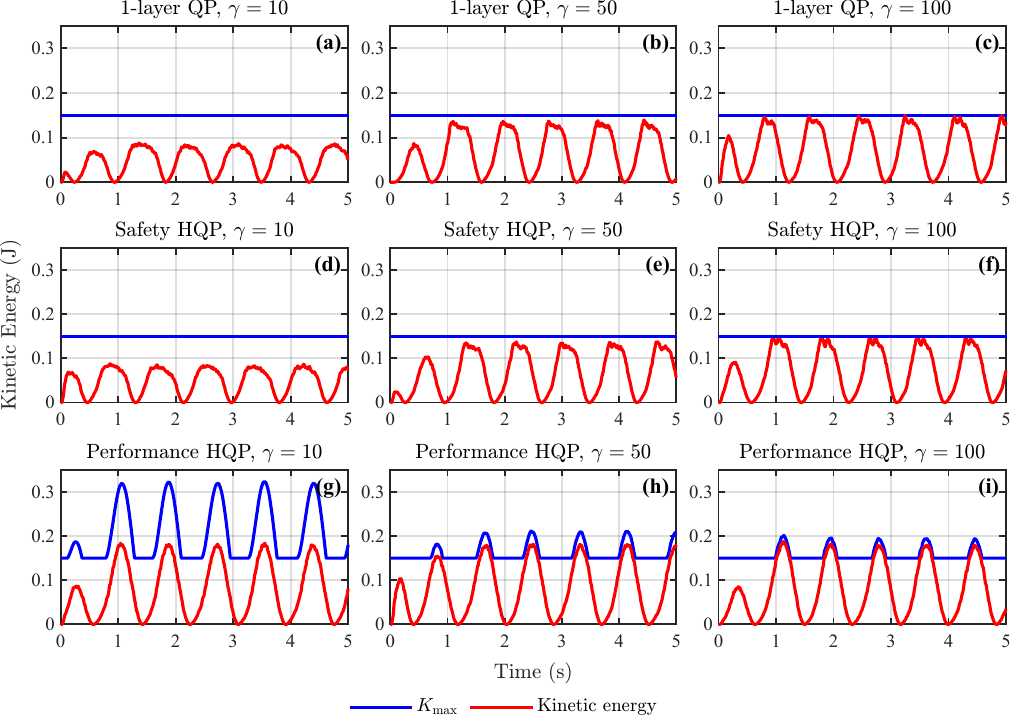}
    \caption{Experiment 2 (Sinusoidal external force): total kinetic energy.}
    \label{fig:Experiment2_1}
\end{figure}

\begin{figure}[!t]
    \centering
    \includegraphics[width=\columnwidth]{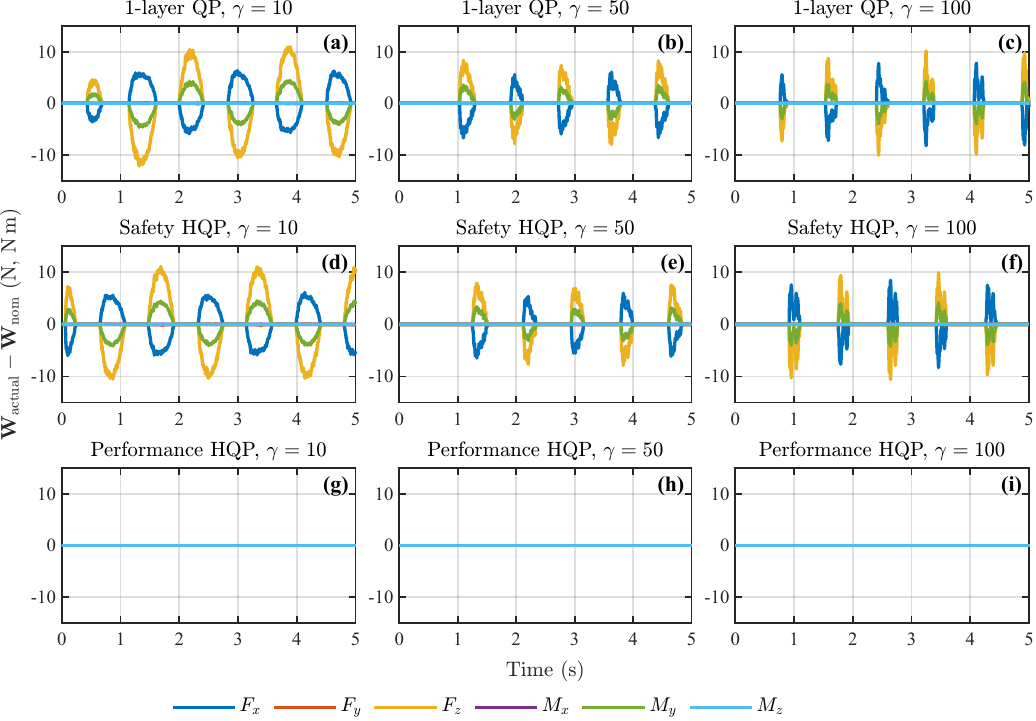}
    \caption{Experiment 2 (Sinusoidal external force): Cartesian wrench deviation.}
    \label{fig:Experiment2_2}
\end{figure}

\begin{figure}[!t]
    \centering
    \includegraphics[width=\columnwidth]{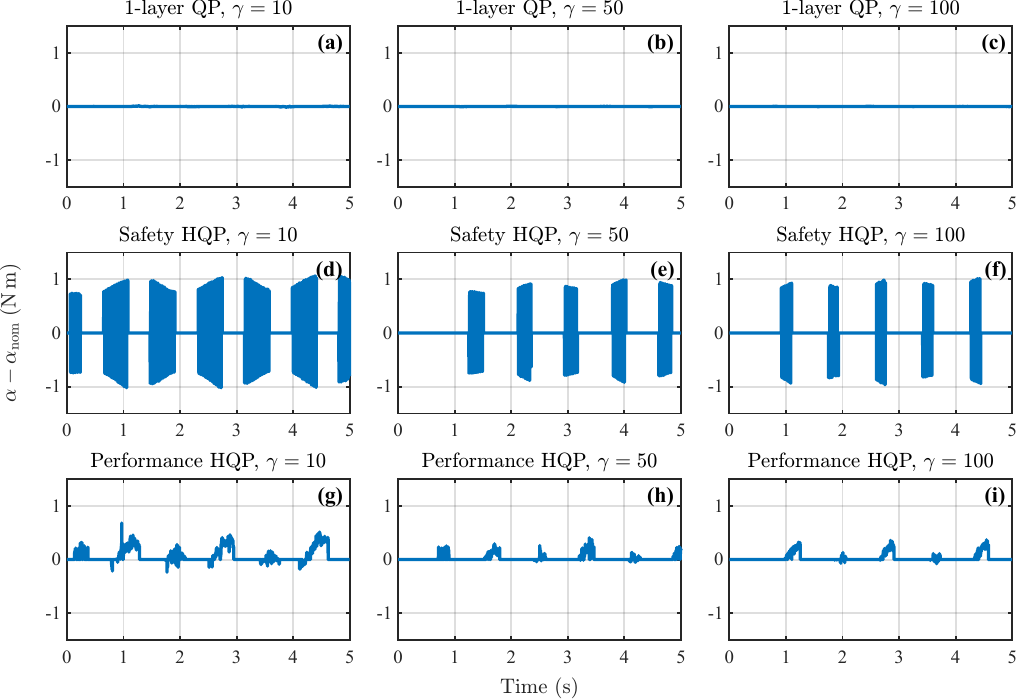}
    \caption{Experiment 2 (Sinusoidal external force): null-space component.}
    \label{fig:Experiment2_3}
\end{figure}

To further evaluate the CBF-HQP performance under a continuous energy injection,
a time-varying external force was applied along the $z$ direction:
\[
F_z(t) = 25 \sin(2\pi \cdot 0.6\, t)\ \mathrm{N}.
\]
This external force periodically injected energy into the system with continuous excitations rather than a single transient burst, as in Experiment~1.

To demonstrate the flexibility of the proposed framework, an additional set of comparative experiments was conducted. In Experiment~1, the HQP with performance-oriented prioritization was evaluated. Here, an alternative configuration was considered, in which safety (the kinetic-energy constraint) was assigned the highest priority.

In Figs.~\ref{fig:Experiment2_1}--\ref{fig:Experiment2_3}, subplots (a--c) are of the single-layer QP baseline method, (d--f) correspond to the safety-priority HQP, and (g--i) correspond to the performance-priority HQP.

Figures~\ref{fig:Experiment2_1} and \ref{fig:Experiment2_2} illustrate the kinetic energy evolution and Cartesian wrench deviation under different values of $\gamma$. When safety was enforced as the highest priority, the bound on kinetic energy was never violated to preserve Cartesian-space performance. In contrast, the performance-priority HQP better preserved the nominal Cartesian behavior while allowing limited energy bound relaxation.

Figure~\ref{fig:Experiment2_3} further confirms the hierarchical behavior in continuous pHRI. While the single-layer formulation kept $\alpha - \alpha_{\mathrm{nom}}$ near zero, the proposed HQP exhibited clear null-space adjustments, indicating strict priority enforcement.

These observations were consistent with Experiment~1 and demonstrated that the proposed HQP achieved prioritized control under sustained human–robot interaction, while providing flexible task prioritization to accommodate different safety and performance requirements.

\section{Discussion}
In this work, we considered the kinetic energy of the entire robot arm. It would be interesting to investigate the decomposition of the entire kinetic energy into the Cartesian-space kinetic energy and the nullspace energy, thereby splitting this safety task into two safety tasks to achieve fine-grained control of the kinetic energy through the CBF-HQP framework.

In practice, it is important to ensure that $S_0$ is nonempty at the beginning of the HQP. This can be achieved by providing the human operator with feedback or guidance on how to avoid this extreme situation, for instance, through a VR headset or other modalities.  

In our examples, the number of QP problems was not large. In case the number is large for a large-scale CBF-HQP, existing algorithms can be used to accelerate the calculation efficiency of solving the sequence of QPs \cite{de2010feature, mansard2012dedicated, escande2014hierarchical}.

\vspace{-0.1cm}
\section{Conclusions}
In this paper, a CBF-based hierarchical quadratic programming framework was proposed to manage multiple performance and safety tasks, including not only kinematics-based safety tasks but also energy-based safety tasks, i.e., kinetic energy limiting, in a flexible manner in the context of continuous pHRI. The effectiveness and flexibility of the proposed method were experimentally validated on a real redundant robot, and the results showed that the approach can balance safety and performance in pHRI depending on different application requirements.


%%%%%%%%%%%%%%%%%%%%%%%%%%%%%%%%%%%%%%%%%%%%%%%%%%%%%%%%%%%%%%%%%%%%%%%%%%%%%%%%

%\vspace{0mm}
\bibliographystyle{IEEEtran}	
\bibliography{IEEEabrv,references}

\end{document}